\documentclass[runningheads]{llncs}
\usepackage{graphicx}
\usepackage[utf8]{inputenc}
\usepackage[T1]{fontenc}
\usepackage{hyphenat}
\usepackage{amssymb}
\usepackage[english]{babel}
\usepackage{multicol}
\usepackage{multirow}
\usepackage{bm}
\usepackage[misc,geometry]{ifsym}

\usepackage{floatrow}

\usepackage{amsmath}

\usepackage{siunitx}

\usepackage{float}
\floatstyle{plaintop}
\restylefloat{table}

\usepackage{caption} 
\usepackage{blindtext,letltxmacro,xcolor,xparse}
\usepackage{lipsum}% for simulating real text

\LetLtxMacro{\blindtextblindtext}{\blindtext}
\LetLtxMacro{\blindtextBlindtext}{\Blindtext}

\RenewDocumentCommand{\blindtext}{O{\value{blindtext}}}{%
  \begingroup\color{gray}\blindtextblindtext[#1]\endgroup
}
\RenewDocumentCommand{\Blindtext}{O{\value{blindtext}}O{\value{Blindtext}}}{%
  \begingroup\color{gray}\blindtextBlindtext[#1][#2]\endgroup
}

\def\BibTeX{{\rm B\kern-.05em{\sc i\kern-.025em b}\kern-.08em
    T\kern-.1667em\lower.7ex\hbox{E}\kern-.125emX}}
    
\usepackage[skip=0.5\baselineskip]{caption}
\newcommand{\bftab}{\fontseries{b}\selectfont}

\makeatletter
\DeclareRobustCommand{\e}[1]{%
  \leavevmode
  \vbox{\offinterlineskip
    \check@mathfonts
    \ialign{%
      ##\cr
      \hidewidth\fontsize{\numexpr\ssf@size/2}\z@\bfseries\itshape e\hidewidth\cr
      \noalign{\kern.2ex}
      #1\cr
    }%
  }%
}
\makeatother

\newfloatcommand{capbtabbox}{table}[\captop][\FBwidth]

\begin{document}

\title{TS-Net: OCR Trained to Switch Between Text Transcription Styles}

\titlerunning{TS-Net: OCR Trained to Switch Between Text Transcription Styles}

\author{Jan Kohút (\Letter) \orcidID{0000-0003-0774-8903} \and
Michal Hradiš\orcidID{0000-0002-6364-129X}}

\authorrunning{J. Kohút et al.}

\institute{Faculty of Information Technology, Brno University of Technology, Brno, Czech~Republic \\
\email{ikohut@fit.vutbr.cz}, \email{ihradis@fit.vutbr.cz}}

%\author{First Author\orcidID{0000-1111-2222-3333} \and
%Second Author\orcidID{1111-2222-3333-4444} \and
%Third Author\orcidID{2222--3333-4444-5555}}

\maketitle              % typeset the header of the contribution
\begin{abstract}
Users of OCR systems, from different institutions and scientific disciplines, prefer and produce different transcription styles.
This presents a problem for training of consistent text recognition neural networks on real-world data. 
We propose to extend existing text recognition networks with a Transcription Style Block (TSB) which can learn from data to switch between multiple transcription styles without any explicit knowledge of transcription rules. 
TSB is an adaptive instance normalization conditioned by identifiers representing consistently transcribed documents (e.g. single document, documents by a single transcriber, or an institution). 
We show that TSB is able to learn completely different transcription styles in controlled experiments on artificial data, it improves text recognition accuracy on large-scale real-world data, and it learns semantically meaningful transcription style embeddings.
We also show how TSB can efficiently adapt to transcription styles of new documents from transcriptions of only a few text lines.

\end{abstract}
\begin{figure}[b]
    \centering
    \includegraphics[width=\linewidth, trim=0mm 105mm 0mm 0mm, clip]{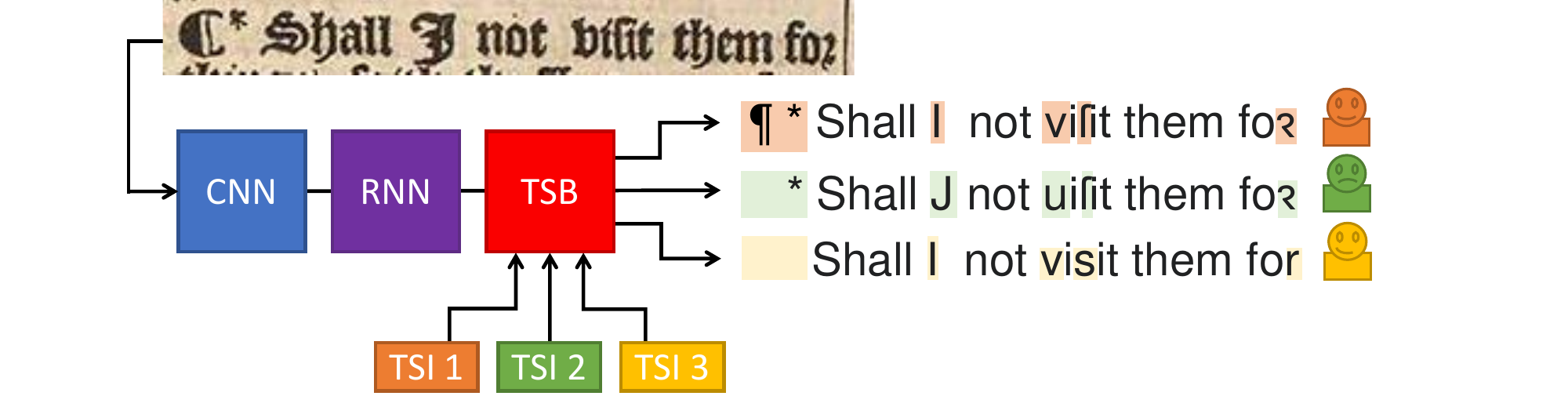}
    \caption{Our proposed Transcription Style Block (TSB) learns to switch between various transcription styles based on transcription style identifiers (TSI).}
    \label{fig:introduction:inconsistent_transcriptions}
\end{figure}

\section{Introduction}

When multiple users transcribe the same document the results might differ. 
The difference might not be given only by errors, but also by the diversity in transcription styles e.g., a historic character called \textquotesingle{}long s\textquotesingle{} can be transcribed with the respective UTF-8 symbol or with the standard \textquotesingle{}s\textquotesingle{}.
As it is not desirable to enforce a specific transcription style on transcribers, especially when it comes to open online systems, the presence of inconsistent transcriptions is inevitable.

The obvious solutions would be to unite the transcription styles of the transcribers or define mappings between the styles. 
Uniting transcription styles might be realistic for a small and closed group of transcribers.
The second solution would require knowledge about the transcription styles in a form of translation tables, which is usually not available. 
If we had a reliable neural network, we could derive the tables in a semi-automatic manner. 
However, such a neural network would require a large amount of consistently transcribed data, resulting in a chicken and egg problem. 
Moreover, a transcription style mapping with $1$ to $n$ character relationships cannot be unambiguously represented. 
A common disadvantage of both options is the inability to automatically preserve personalized transcription styles.

OCR neural networks~\cite{SceneCTC2015,PuigcerverCTC2017,GatedCTC2017,ImprovingCTC2018} trained with CTC loss function~\cite{CTC2006} learn to average probabilities of inconsistently transcribed characters if the transcription style can not be estimated from the input image, resulting in mostly random and independent transcription choices for such characters.
The Seq2Seq approaches~\cite{EvaluatingSeq2Seq2019,EfficientSeq2Seq2018} can produce locally consistent transcriptions because of the autoregressive decoding process, but the choice of a transcription style may be arbitrary. 

Figure~\ref{fig:introduction:inconsistent_transcriptions} shows our proposed solution. 
Apart from the input, the neural network takes a transcription style identifier (TSI) and outputs a transcription in the respective style. 
A TSI identifies data with the same transcription style, it might be an identifier of a  transcriber, of a document, or of any other transcription consistent entity. 
There is no need for any explicit knowledge (translation tables/rules) about the transcription styles (TS).
TS are automatically learned by Transcription Style Block (TSB), which can be added near the end of any standard neural network (in our case convolutional layers (CNN) followed by BLSTM~\cite{LSTM1997} layers (RNN) with the CTC loss function). 
TSB is an adaptive instance normalization layer with an auxiliary network, which learns to generate scales and offsets for each SI.
It is trained jointly with the rest of the neural network.

To test if TSB is able to handle extreme scenarios, we evaluated our system on two datasets with synthetic TS.
A synthetic TS is a full permutation of the character set. 
The first dataset consists of synthetic text line images with random text, the second is the IMPACT dataset~\cite{IMPACT2013}.
To show practical benefits, we experimented with documents from Deutsches Textarchiv (DTA) where the transcribers are inconsistent among a handful of characters.
We show that based on text line images, transcriptions, and TSI, the neural network is able to learn TS automatically.
The adaptation to a new TS is fast and can be done with a couple of transcribed informative text line images.

\section{Related work}

To our knowledge, no existing text recognition approach can automatically handle multiple transcription styles both during training and inference.
Usually, there is an effort to avoid any inconsistencies in transcriptions from the start by introducing a transcription policy and resolving conflicts manually~\cite{IMPACT2013,Bentham2018}.

As our approach is based on adaptation to transcription style, it is closely related to the neural networks adaptation field. 
In text recognition, several authors explored various techniques for domain adaptation, transfer learning, and similar~\cite{Seq2SeqAdaptation2019,UnsupervisedSyntheticToRealAdaptation2019,TextAdaptation2019}, but these techniques produce a unique model for each target domain and the individual domains have to be clearly defined. 
The idea of model adaptation was more extensively explored in speech recognition~\cite{SpeechAdaptationOverview2020}, mainly in the form of structured transforms approaches, which allow models to efficiently adapt to a wide range of domains or even handle multiple data domains simultaneously. 

Model adaptation in speech recognition targets mostly the variability of individual speakers and communication channels. 
The main idea of structured transforms approaches is to adapt a subset of network parameters to speaker identity (speaker-dependent, SD) and keep the rest of the network speaker-independent (SI).
The published methods adapt input layer (linear input network, LIN~\cite{LIN1995}), hidden layer (linear hidden network, LHN~\cite{LHN2007}), or output layer (linear output network, LON~\cite{LON2010}). 
The main disadvantage of these methods is the large number of adapted parameters for each speaker. 
Adaptation is therefore slow and it is prone to overfitting without strong regularization. 

Many methods strive to decrease the number of speaker-dependent parameters to reduce overfitting and improve adaptation speed.
Learning Hidden Unit Contributions (LHUC)~\cite{LHUC2016} adds an SD scale parameter after every neuron (kernel). 
Zhang et al.~\cite{ParametrisedActivation2015} parametrized activation functions (ReLU and Sigmoid) while making some of these parameters SD. 
Wang et al.~\cite{BNOfflineAdaptation2017} and Mana et al.~\cite{BNOnlineAdaptation2019} repurposed scales and offsets of batch normalization as SD parameters.  
Zhao et al.~\cite{LRPD2016,eLRPD2017} found that most of the information in SD FC layers is stored in diagonals of the weight matrices and they proposed Low-Rank Plus Diagonal (LRPD, eLRPD)  approaches which decompose (factorize) the original weight matrix of an FC layer into its diagonal and several smaller matrices.
Factorized Hidden Layer (FHL)~\cite{FHL2016} offers a similar solution. 

Another way to restrict the number of parameters is to use an SI auxiliary network that generates SD parameters based on a small SD input.
The SD input representing the speaker can be an i-vector, learned speaker embedding, or similar features. 
Delcroix et al.~\cite{ContextAdaptive2018} aggregate outputs of several network branches using weights computed from the SD input. 
The assumption is that the branches would learn to specialize in different types of speakers.
Cui et al.~\cite{EmbeddingBased2017} used an auxiliary network to generate SD scales and offsets applied to hidden activations. 

Some methods do not require explicit information about speaker identity and learn to extract such SD information from a longer utterance and distribute it locally. 
L. Sar{\i} et al.~\cite{SpeakerOffset2019} trained an auxiliary network to predict SD single layer local activation offsets by pulling information from an entire utterance.
Similarly, Kim et al.~\cite{DynamicLayerNormalization2017} predict activation offsets and scales of multiple layers. 
Xie et al.~\cite{LHUCOnline2019} use an auxiliary network to generate the LHUC parameters in an online fashion based on acoustic features. 

When the number of SD parameters is relatively small, the structure transforms approaches allow efficient network adaptation which is resistant to overfitting and requires only a small amount of adaptation data. 
As the speaker correlated information generally vanishes in deeper layers of a network~\cite{SI2012}, the SD parameters are usually located in the early layers of the network (in the acoustic part).

Our approach shares many ideas with the existing structured transforms approaches as the Transcription Style Block (TSB) uses learned domain embeddings transformed by an auxiliary network to scale and offset activations. 
However, this adaptation is done near the network output as it aims to change the output encoding, not to adapt to different styles of inputs. 

The architecture of TSB is inspired by style transfer approaches and style-dependent Generative Adversarial Networks. 
These methods often use an adaptive instance normalization (AdaIN) to broadcast information about the desired output style across a whole image~\cite{CIN2016,AdaIN2017,AuxStyle2017,StyleGAN2018}.
The published results demonstrate that this network architecture can efficiently generate a large range of output styles while keeping the style consistent across the whole output.

\section{Transcription Style Block}

\begin{figure}[t]
    \centering
    \includegraphics[width=\linewidth, trim=0mm 0mm 0mm 0mm, clip]{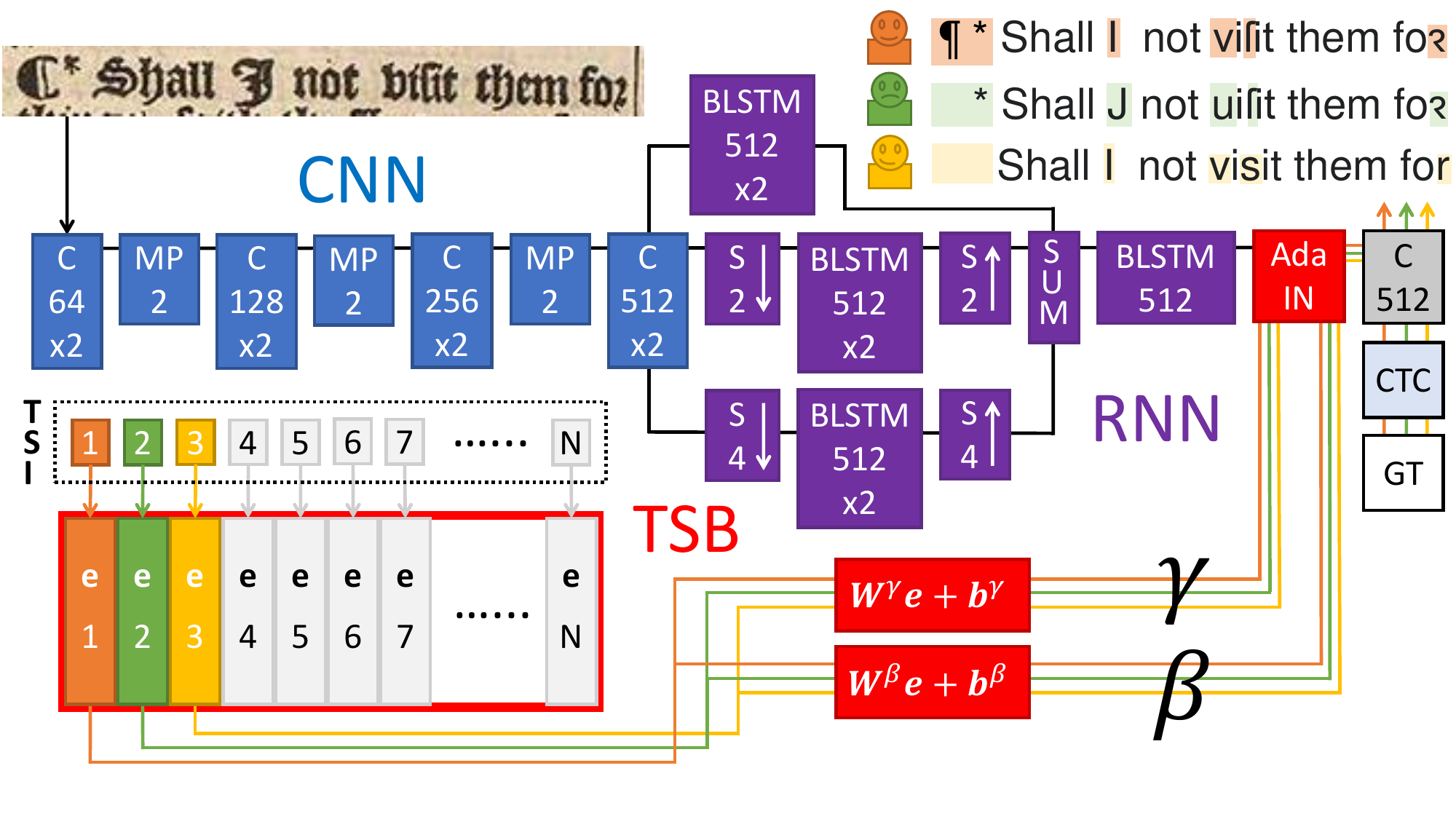}
    \caption{Our proposed neural network takes a text line image (top left) and a transcription style identifier (TSI) and outputs a transcription in the respective style (top right). 
    It consists of three main parts, convolutional part (CNN, blue), recurrent part (RNN, purple), and the Transcription Style Block (TSB, red).
    Each TSI is represented by a transcription style embedding $\bm{e}$. 
    Affine transformations of an $\bm{e}$ results in the scales and offsets, $\bm{\gamma}$ and $\bm{\beta}$, for the adaptive instance normalization (AdaIN).}
    \label{fig:introduction:multi_transcription_styles_model}
\end{figure}

Our neural network architecture (see Figure~\ref{fig:introduction:multi_transcription_styles_model}) is similar to the current state of the art text recognition networks~\cite{SceneCTC2015,PuigcerverCTC2017,GatedCTC2017,ImprovingCTC2018} which use CTC loss~\cite{CTC2006}. 
It consists of three main parts, convolutional part (CNN, blue), recurrent part (RNN, purple), and the Transcription Style Block (TSB, red). 
TSB allows the network to change the transcription style of the output (orange, green, yellow, etc.) based on the style transcription identifier (TSI).
The transcription style switching is realized by adaptive instance normalization (AdaIN)~\cite{AdaIN2017}: 
\begin{equation}
\text{AdaIN}(\mathbf{X}_c, \gamma_c, \beta_c) = \gamma_c\bigg(\frac{\mathbf{X}_{c} - \mu(\mathbf{X}_{c})}{\sigma(\mathbf{X}_{c}) + \epsilon}\bigg) + \beta_c,
\end{equation}
where $\mu(\mathbf{X}_{c})$, $\sigma(\mathbf{X}_{c})$ are mean and standard deviation of a channel $c$ in a activation map $\mathbf{X}$  ($\epsilon$ is a small positive number). 
A transcription style is changed by scaling and offsetting the normalized channels by $\gamma_c$ and $\beta_c$, respectively. 
Vectors of scales and offsets $\bm{\gamma}$, $\bm{\beta}$ are given by affine transformations of a transcription style embedding $\bm{e}$: 
\begin{equation}
     \bm{\gamma} = \bm{W}^\gamma \bm{e} + \bm{b}^\gamma, \bm{\beta} = \mathbf{W}^\beta \bm{e} + \bm{b}^\beta, 
\end{equation}
where $\bm{W}^\gamma$, $\bm{W}^\beta$ are the transformation matrices and $\bm{b}^\gamma$, $\bm{b}^\beta$ biases.
Every TSI is therefore represented by one transcription style embedding $\bm{e}$.
The output of the recurrent part is transcription style independent feature vectors that are turned into the transcription style dependent by the TSB. 
We decided to place the TSB near the end of the architecture as there should be no need for transcription style dependent processing in the visual and context part.

\paragraph{Training.} 
Our datasets consist of triplets of text line images, transcriptions, and TSI, which are integer numbers representing transcription consistent entities. 
The whole neural network is trained jointly with Adam optimizer \cite{Adam2017}. 
The only transcription style dependant parameters are the embeddings, the rest of the parameters, including $\bm{W}^\gamma$, $\bm{W}^\beta$, $\bm{b}^\gamma$, and $\bm{b}^\beta$ are updated with each training sample. 
We initialize each embedding from standard normal distribution $\mathcal{N}(0,1)$. 
The transformation matrices $\bm{W}^\gamma$, $\bm{W}^\beta$ are initialized from uniform distribution $\mathcal{U}(-k,k)$, where $k$ is a small positive number (more details in section~\ref{sec:synthetic_transcription_styles_experiments}, Table~\ref{tab:experiments:random:IMPACT:W_init}). 
The biases $\bm{b}^\gamma$ are set to ones, the biases $\bm{b}^\beta$ are set to zeros. 

We use data augmentation including color changes, geometric deformations, and local masking which replaces the input text line image with noise at random locations and with random width limited to roughly two text characters. 
The idea of masking is to force the network to model language and not to rely only on visual cues. 
Further details about training and datasets are given in sections~\ref{sec:synthetic_transcription_styles_experiments} and~\ref{sec:real_transcription_styles_experiments}. 

\paragraph{Architecture details.}
The convolutional part (blue) is similar to the first 4 blocks of VGG 16~\cite{VGG2015} with the exception that only the first two pooling layers reduce horizontal resolution and Filter Response Normalization (FRN) \cite{FRN2019} is placed at the output of each block. 
It consists of four convolutional blocks (C) interlaid with max pooling layers (MP) with kernel and step size of 2.
Each convolutional block consists of two (x2) convolutional layers with 64, 128, 256, and 512 output channels, respectively. 
The last convolutional layer aggregates all remaining vertical pixels by setting the kernel height equal to the height of its input tensor.

The subsequent recurrent part (purple) consists mainly of BLSTM layers~\cite{LSTM1997}. 
It processes the input by three parallel BLSTM blocks (each block has two BLSTM layers). 
The first block processes directly the convolutional output, the second and third work at twice (S2$\downarrow$) and four (S4$\downarrow$) times reduced horizontal resolution, respectively. 
Outputs of all three branches are converted back to the same resolution by nearest neighbor upsampling and are summed together and processed by the final BLSTM layer.
The reason for this multi-scale processing is to enlarge the receptive field (context) of the output hidden features.

\section{Synthetic Transcription Styles Experiments}
\label{sec:synthetic_transcription_styles_experiments}

To test if the proposed TSB is indeed able to learn different transcription styles by itself and that this functionality is not learned by the whole network relying on correlations between the transcription style and visual inputs, we evaluated the whole network in extreme artificial scenarios. 
These experiments also give information about the limits and scalability of this approach.
We evaluated our system on two datasets (SYNTHETIC and IMPACT) with 10 synthetic transcription styles (TS). 
A synthetic TS is a full permutation of the character set, e.g. if (a, b, c) is the visual character set then the TS (c, a, b) transcribes the character \textquotesingle a\textquotesingle{} as the character \textquotesingle c\textquotesingle{}, etc. 
In real-life datasets, the difference between TS would be limited to a couple of characters. 
To test the robustness of the system, we represented each TS by multiple transcription style identifiers TSI (up to 10k) and we investigated the correlation between the respective learned transcription style embeddings and TS.

In all following experiments, the transcription style agnostic baseline network replaces TSB with FRN normalization. Also as we show the results for augmented training data (TRN), the results are worse than for the testing data (TST). 

\begin{figure}[t]
    \centering
    \includegraphics[width=\linewidth, trim=60mm 80mm 60mm 67mm, clip]{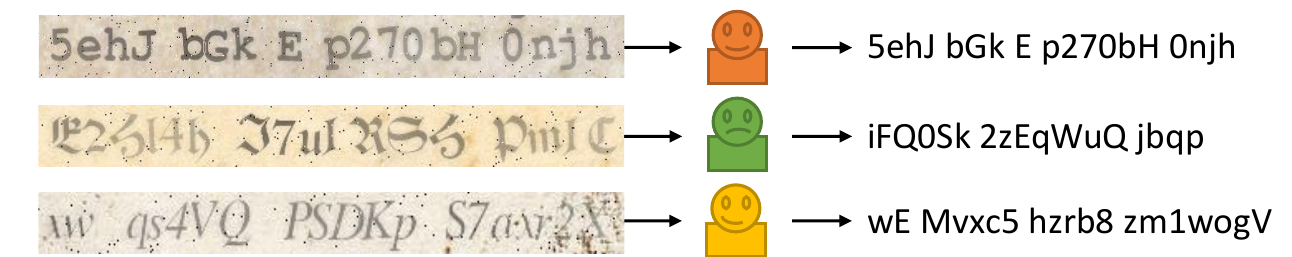}
    \caption{Synthetic text line images with random text and random transcription styles.}
    \label{fig:experiments:random:synthetic_random_transcription_styles}
\end{figure}

\paragraph{Datasets.}
The SYNTHETIC dataset contains generated images of random text lines rendered with various fonts, including historic fonts (see Figure~\ref{fig:experiments:random:synthetic_random_transcription_styles}). 
The images were degraded by simulated whitening, noise, stains, and other imperfections, and slight geometric deformations were introduced in the text (character warping, rotation, and translation).
It should be noted that the visual complexity of this dataset is mostly irrelevant because the experiments focus on TS.
We generated 130k text lines for each of 10 synthetic TS.
All the 10 TS are unique mappings for all characters, except for the white space. 
The size of the character set is 67. 

The IMPACT dataset~\cite{IMPACT2013} contains page scans and manually corrected transcriptions collected as a representative sample of historical document collections from European libraries. 
The documents contain 9 languages and 10 font families. 
We randomly sampled 1.3M text line images from the original collection with ParseNet~\cite{Layout2021} and we translated their transcriptions into additional nine synthetic TS. 
Thus, each text line has ten transcription versions where one is the original one. 
The size of the character set is 399.

\begin{table}[t]
\caption{CER (in \%) on the SYNTHETIC and the IMPACT datasets with 10 synthetic transcription styles and for various embedding dimensions (ED). 
NA is the baseline network trained and tested only on a single TS.}
\label{tab:experiments:synthetic:ED}
\centering
{\renewcommand{\arraystretch}{1.1}%
\tabcolsep=0.2cm
\begin{tabular}{c | S[table-format=2.2] S[table-format=1.2] S[table-format=1.2] S[table-format=1.2] S[table-format=1.2] S[table-format=1.2] S[table-format=1.2] | S[table-format=1.2]}
{ED}  & {1} & {2} & {4} & {8} & {16} & {32} & {64} & {NA} \\ \hline\hline 
{TST SYNTHETIC} & 2.95 & 0.58 & 0.66 & 0.50 & 0.67 & \bftab 0.32 & 0.38 & 0.35 \\
{TRN SYNTHETIC} & 5.72 & 1.70 & 1.81 & 1.59 & 1.77 & \bftab 1.06 & 1.29 & 0.91 \\ \hline
{TST IMPACT} & 10.87 & 2.57 & 1.13 & 1.09 & 1.09 & \bftab 0.89 & 0.99 & 0.74 \\
{TRN IMPACT} & 12.04 & 3.46 & 2.00 & 1.94 & 1.93 & \bftab 1.53 & 1.71 & 1.27 \\
\end{tabular}}
\end{table}

\paragraph{Transcription style embedding dimension.}
The choice of the embedding dimension (ED) influences the ability of the network to transcribe the input in various TS. 
Fewer tunable parameters should provide higher overfitting resistance but may not be able to capture all the TS differences.
The goal is to find the smallest ED, which will not degrade the text recognition accuracy.  
In this experiment, each TS is uniformly represented by multiple TSI (TSI per TS). 
Table~\ref{tab:experiments:synthetic:ED} shows the test and the train character error rates (CER) for various ED. 
10 TSI per TS was used for the SYNTHETIC dataset and 100 TSI per TS for the IMPACT dataset (as the IMPACT dataset contains 10 times more data, each embedding is trained on approximately 13k text line images). The networks were trained for 66k iterations. 
To estimate an upper bound on transcription accuracy, the baseline network (NA) was trained on a single TS. 

The ED 32 brought the best results. 
Surprisingly, the network with ED 1 is already able to switch the transcription style to a limited degree, especially considering that all AdaIN scales and all offsets in this case differ just by a scale constant.
The results of the ED 32 and these of the baseline network NA are comparable, which means that the TSB learned to switch between the transcription styles without lowering the performance of the text recognition. 

\begin{table}[t]
\RawFloats
\parbox[t][][t]{.45\linewidth}{
\centering
{\renewcommand{\arraystretch}{1.1}%
\tabcolsep=0.15cm
\caption{CER (in \%) on the SYNTHETIC dataset for various TSI per TS, ED 32.}
\label{tab:experiments:random:synthetic:TSIperTS}
\begin{tabular}{c | c c c}
TSI per TS & 10 & 100 & 1000 \\ \hline\hline 
TST & 0.32 & 0.35 & 0.29 \\
TRN & 1.06 & 1.13 & 1.01 \\
lines per E & 13k & 1.3k & 130 \\
TRN iterations & 66k & 132k & 330k \\
\end{tabular}}
}
\hfill
\parbox[t][][t]{.45\linewidth}{
\centering
{\renewcommand{\arraystretch}{1.1}%
\tabcolsep=0.15cm
\caption{CER (in \%) on the IMPACT dataset for various TSI per TS, ED 32.}
\label{tab:experiments:random:IMPACT:TSIperTS}
\begin{tabular}{c | c c c}
TSI per TS & 100 & 1000 & 10000 \\ \hline\hline 
TST & 0.89 & 1.40 & 1.40 \\
TRN & 1.53 & 2.50 & 2.00 \\
lines per E & 13k & 1.3k & 130 \\
TRN iterations & 66k & 100k & 378k \\
\end{tabular}}
}
\end{table}

\begin{figure}
\TopFloatBoxes
\begin{floatrow}
\ffigbox[4.6cm]{
  \centering
  \includegraphics[width=0.88\linewidth, trim=0mm 0mm 0mm 0mm, clip]{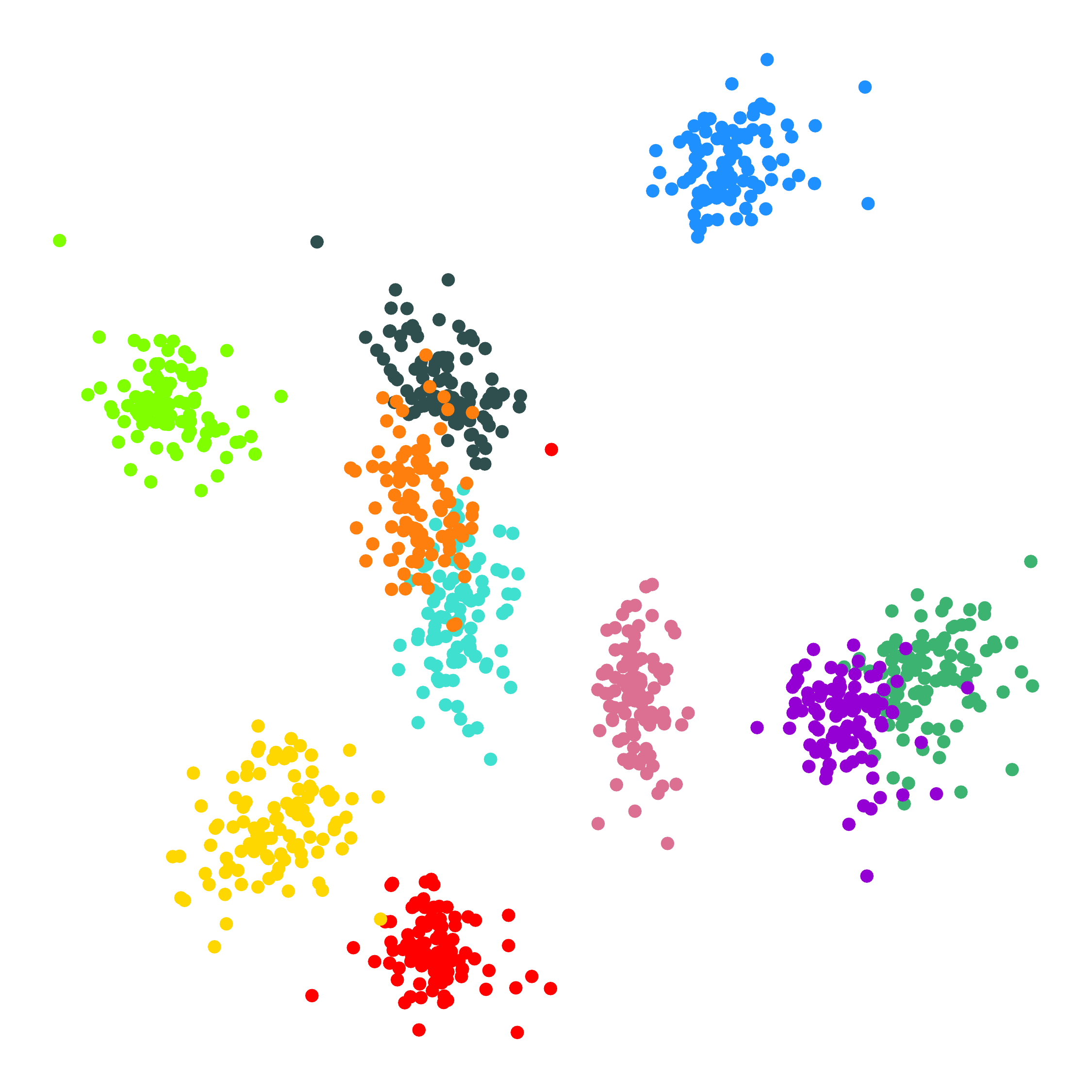}
}{
  \caption{2D projection (PCA) of the AdaIN scales and offsets for all embeddings learned on the SYNTHETIC dataset with 100 TSI per TS.}%
  \label{fig:experiments:random:synthetic_PCA}
}
\capbtabbox{
  {\renewcommand{\arraystretch}{1.1}
   \tabcolsep=0.1cm
   \begin{tabular}{c | c c c c c c}
     $k$ & 0.03 & 0.075 & 0.15 & 0.3 & 0.6 & 1.2  \\ \hline\hline 
     TST & 0.91 & 0.90 & 0.89 & 1.01 & 0.96 & 1.48 \\
     TRN & 1.55 & 1.57 & 1.53 & 1.70 & 1.70 & 2.59 \\
    \end{tabular}}
}
{
  \caption{CER (in \%) for various initialization energy of $\bm{W}^{\gamma}$, $\bm{W}^{\beta}$ matrices. The weights were initialized by sampling from uniform distribution $\mathcal{U}(-k,k)$ .}
  \label{tab:experiments:random:IMPACT:W_init}
}
\end{floatrow}
\end{figure}

\paragraph{Number of transcription style identifiers per transcription style.} 
The goal of the following experiment is to show if the network is capable to learn TS from a relatively small number of lines, and also if the network is able to handle a large amount of TSI. 
We can explore this by increasing the number of TSI per TS. 
We note that the network is not aware of which TSI represent the same TS in any way and because all embeddings are randomly initialized, it must learn the TS of TSI only from the respective text line images and transcriptions. Tables~\ref{tab:experiments:random:synthetic:TSIperTS} and~\ref{tab:experiments:random:IMPACT:TSIperTS} show that the proposed approach scales well with the number of TSI and the final achieved transcription accuracy does not seem to depend on the number of TSI. 
However, the network needs more training iterations to converge with an increasing number of TSI. Consider that for the highest explored number of TSI, only 130 text line images were available for each TSI.

Figure~\ref{fig:experiments:random:synthetic_PCA} shows 2D projection (PCA) of the AdaIN scales and offsets for all embeddings learned on the SYNTHETIC dataset with 100 TSI per TS. 
The projections of different TS are well separated and it is apparent that the network learned a semantically meaningful embedding space. 
We observed that the separation increases during training proportionally to the error rate reduction.

Additionally we experimented with initialization of the matrices $\bm{W}^{\gamma}$ and $\bm{W}^{\beta}$. 
We trained the network on the IMPACT dataset for 66k iterations, with 100 TSI per TS and ED 32.
Table~\ref{tab:experiments:random:IMPACT:W_init} shows that the network is very robust to the initial weight distribution energy represented as $k$, and that it converges to similar error rates until the energy exceeds a specific threshold.
We observed slightly faster convergence for lower initial energies.

\section{Real Transcription Styles Experiments}
\label{sec:real_transcription_styles_experiments}

To demonstrate the benefits of our system in a practical application, we experimented with a collection of documents from Deutsches Textarchiv (DTA)\footnote{https://www.deutschestextarchiv.de}.
DTA collects German printed documents from various fields, genres, and time periods. 
Most documents date from the early 16th to the early 20th century and are typeset mostly in Fraktur fonts. 
We chose this collection because it contains several transcription styles related to archaic character variants.
For example, a character called \textquotesingle{}long s\textquotesingle{} is often transcribed as the standard \textquotesingle{}s\textquotesingle{}, also the German umlauted vowels in old notation are sometimes transcribed with their modern versions. 

\paragraph{DTA dataset.} 
Our DTA dataset is a subset of the original DTA collection. 
We represent a document by a single TSI, as we assume that each document was transcribed by a single organization using a consistent transcription style, but we have no information on how or where the documents were transcribed. 
We randomly sampled 1.08M text lines from 2445 randomly picked DTA documents with ParseNet~\cite{Layout2021}. 
The number of lines per document ranges from 200 to 3029.
The size of the character set is 271.  
The test set (TST) contains 12.5k random samples, the train set (TRN) contains the rest.
We also created two additional held-out test sets of documents that are not present in the TRN/TST partitions. 
These two sets allowed us to assess the behavior of the model on unknown documents.
The first set (TST-HO) contains 12.5k lines from 7905 documents.
The second set (TST-C) contains only 1355 lines, each line is from a different document of the TST-HO set. 
We transcribed TST-C consistently and accurately with the original graphemes.

\begin{table}[t]
\caption{CER (in \%) on the DTA dataset for various ED and the baseline network NA, 200k training iterations. TST-STSI refers to the TST set with randomly shuffled TSI.}\label{tab:experiments:real:basic}
\centering
{\renewcommand{\arraystretch}{1.1}%
\tabcolsep=0.2cm
\begin{tabular}{c | c c c c c c c | c}
ED & 1 & 2 & 4 & 8 & 16 & 32 & 64 & NA \\ \hline\hline 
TST & 0.62 & 0.59 & 0.58 & 0.54 & 0.52 & \bftab 0.51 & 0.51 & 0.66 \\
TST-STSI & 0.86 & 0.86 & 1.22 & 1.57 & 1.73 & \bftab 1.74 & 1.81 & -  \\
TRN & 0.79 & 0.80 & 0.70 & 0.70 & 0.71 & \bftab 0.72 & 0.70 & 0.79 \\
\end{tabular}}
\end{table}

\paragraph{TSB performance on real data.}
We trained our system with different embedding dimensions (ED), and the baseline network, for 200k iterations.
The results are shown in Table~\ref{tab:experiments:real:basic}. 
In comparison to the baseline network (NA), our system brought 22.7\% relative decrease in character error rate on the TST set and 11.4\% on the augmented TRN set.
There is a noticeable drop in accuracy when we randomly shuffle TSI which means TSB learned to change the transcriptions and did not degrade into standard instance normalization.
The performance for larger ED improves, but gradually worsens when TSI are shuffled. 
This behavior indicates that larger embeddings are better in capturing the transcription styles. 
As the ED 64 did not bring any significant improvement, we chose the ED 32 for the following experiments (it was also the best ED for the experiments in section~\ref{sec:synthetic_transcription_styles_experiments}).

TSB reduces CER on the DTA dataset even if we allow the output to match any transcription style. We verified this by transforming transcriptions of the DTA TST set and the network outputs to a consistent modern style by a hand-designed look up table (reflecting substitution statistics, see Table~\ref{tab:experiments:real:MRM}). In this setting network with TSB (ED 32) provides 0.504\% CER and network without TSB 0.568\% CER.

\paragraph{Analysis of DTA transcription style inconsistencies.}
Characterizing the transcription styles learned by the network is not straightforward.
We analyzed the produced transcriptions, their differences, and relations.
First, we explore which characters are substituted with changing TSI. 
Average statistics of the substations should clearly indicate if the changes are meaningful and correspond to expected transcription styles or if they are random and correspond only to errors or uncertain characters.

We gathered character substitution statistics on held-out documents (TST-HO) by comparing outputs of all TSI pairs using standard string alignment (as used for example in Levenshtein distance). 
We aggregated the statistics over all TSI pairs to identify globally the most frequently substituted character pairs.

\begin{table}[t]
\centering
{\renewcommand{\arraystretch}{1.1}%
\tabcolsep=0.12cm
\begin{tabular}{c | c c c | c c c c c c | c c | c c | c c | c c c c}
  & {$\lceil$} & {s} & {ß} & {$\overset{\scriptscriptstyle\text{e}}{\text{a}}$} & {ä} & {$\overset{\scriptscriptstyle\text{e}}{\text{o}}$} & {ö} & {$\overset{\scriptscriptstyle\text{e}}{\text{u}}$} & {ü} & {I} & {J} & {$\imath$} & {r} & {\textquotesingle} & {\textsf{’}} & {¬} & {=} & {—} & {-} \\ \hline\hline
{\multirow{2}{*}{top 1}} & {s} & {$\lceil$} & {s} & {ä} & {$\overset{\scriptscriptstyle\text{e}}{\text{a}}$} & {ö} & {$\overset{\scriptscriptstyle\text{e}}{\text{o}}$} & {ü} & {$\overset{\scriptscriptstyle\text{e}}{\text{u}}$} & {J} & {I} & {r} & {$\imath$} & {\textsf{’}} & {\textquotesingle} & {-} & {-} & {$\varnothing$} & {¬} \\ 
 & 97 & 97 & 92 & 82 & 77 & 89 & 85 & 83 & 81 & 96 & 97 & 73 & 34 & 93 & 93 & 97 & 84 & 52 & 50 \\ \hline
{\multirow{2}{*}{top 2}} & {$\varnothing$} & {ß} & {z} & {$\overset{\scriptscriptstyle\text{e}}{\text{.}}$} & {$\varnothing$} & {o} & {o} & {$\overset{\scriptscriptstyle\text{e}}{\text{.}}$} & {$\varnothing$} & {$\varnothing$} & {$\varnothing$} & {e} & {$\varnothing$} & {$\varnothing$} & {$\varnothing$} & {$\varnothing$} & {$\varnothing$} & {-} & {$\varnothing$} \\
 & 2 & 2 & 3 & 11 & 11 & 6 & 8 & 9 & 9 & 1 & 1 & 16 & 20 & 4 & 5 & 2 & 15 & 37 & 20 \\ \hline
{\multirow{2}{*}{top 3}} & {f} & {$\varnothing$} & {$\lceil$} & {a} & {a} & {$\overset{\scriptscriptstyle\text{e}}{\text{.}}$} & {$\varnothing$} & {u} & {u} & {1} & {F} & {$\varnothing$} & {t} & {°} & {e} & {.} & {—} & {\textvisiblespace} & {=} \\ 
 & 1 & {<} & 2 & 7 & 10 & 5 & 6 & 5 & 8 & 1 & {<} & 3 & 14 & 1 & {<} & {<} & 2 & 3 & 11 \\ \hline
{\multirow{2}{*}{top 4}} & {l} & {f} & {$\varnothing$} & {á} & {ü} & {$\varnothing$} & {h} & {\r{u}} & {ä} & {l} & {l} & {\&} & {x} & {\textquotedbl} & {\textvisiblespace} & {r} & {\textvisiblespace} & {„} & {—} \\ 
 & {<} & {<} & 2 & {<} & 1 & {<} & {<} & 2 & 1 & 1 & {<} & 2 & 7 & 1 & {<} & {<} & {<} & 2 & 6 \\ \hline
{R} & 12 & 13 & 2 & 10 & 7 & 10 & 9 & 9 & 9 & 17 & 11 & 11 & {<} & 45 & 42 & 85 & 44 & 7 & 6
\end{tabular}}
\caption{The most substituted characters of the learned transcription styles. 
The statistics were gathered on the DTA TST-HO set.
Each column describes how the character in the header is substituted.
The \textquotesingle{}top $n$\textquotesingle{} rows show with which characters it is substituted together with the relative frequency of those substitutions.
R is a ratio of substituted character occurrences.
All numerical values are percentages.
Some characters in the table represent similar historic characters or special substitutions. The mapping is: \textquotesingle{}$\lceil$\textquotesingle{} - long~s;
\textquotesingle{}$\imath$\textquotesingle{} - rotunda~r;
\textquotesingle{}\textvisiblespace{}\textquotesingle{} - white space;
\textquotesingle{}$\overset{\scriptscriptstyle\text{e}}{\text{.}}$\textquotesingle{}  - historic umlaut without the base vowel; 
$\varnothing$ - character deletion;
\textquotesingle{}<\textquotesingle{} stands for negligible values lower than 1.}
\label{tab:experiments:real:MRM}
\end{table}

The most frequently substituted (inconsistent) characters in relative and absolute terms are shown in Table~\ref{tab:experiments:real:MRM}.
The substitutions are mostly unambiguous and match our expectations of possible transcription styles of historic German texts, i.e. substitutions of archaic and modern character variants.
Other substitutions represent more technical aspects of transcription styles such as transcription of different dashes and hyphenation characters.
Perhaps most interestingly, the network learned with some TSI to differentiate between \textquotesingle{}I\textquotesingle{} and \textquotesingle{}J\textquotesingle{} even though these modern characters are represented by the same grapheme in most documents.

As expected, the most inconsistent pair is \textquotesingle{}s\textquotesingle{} and \textquotesingle{}long s\textquotesingle{} ($\lceil$). 
The substitutions of the character \textquotesingle{}sharp s\textquotesingle{} (ß) are meaningful but rare. 
The next group represents inconsistent transcriptions of historic umlauted vowels
\textquotesingle{}$\overset{\scriptscriptstyle\text{e}}{\text{a}}$\textquotesingle{}, 
\textquotesingle{}$\overset{\scriptscriptstyle\text{e}}{\text{o}}$\textquotesingle{}, 
\textquotesingle{}$\overset{\scriptscriptstyle\text{e}}{\text{u}}$\textquotesingle{}. 
As the umlauted vowels in the old notation were transcribed with two separate characters, sometimes the network outputs just the historic umlaut without the base character (represented as \textquotesingle{}$\overset{\scriptscriptstyle\text{e}}{\text{.}}$\textquotesingle{}).
Other significant inconsistent groups are \textquotesingle{}rotunda r\textquotesingle{}($\imath$) and \textquotesingle{}r\textquotesingle{}, single quotes, and various forms of hyphen or other characters that were used for word splitting (\textquotesingle{}¬\textquotesingle{}, \textquotesingle{}=\textquotesingle{}). 
When interpreting the substitution table, keep in mind that the substitutions are not with respect to the real graphemes, instead, they are substitutions between the different learned transcription styles.

\begin{figure}[t]
    \centering
    \includegraphics[width=0.8\textwidth, trim=0mm 0mm 0mm 0mm, clip]{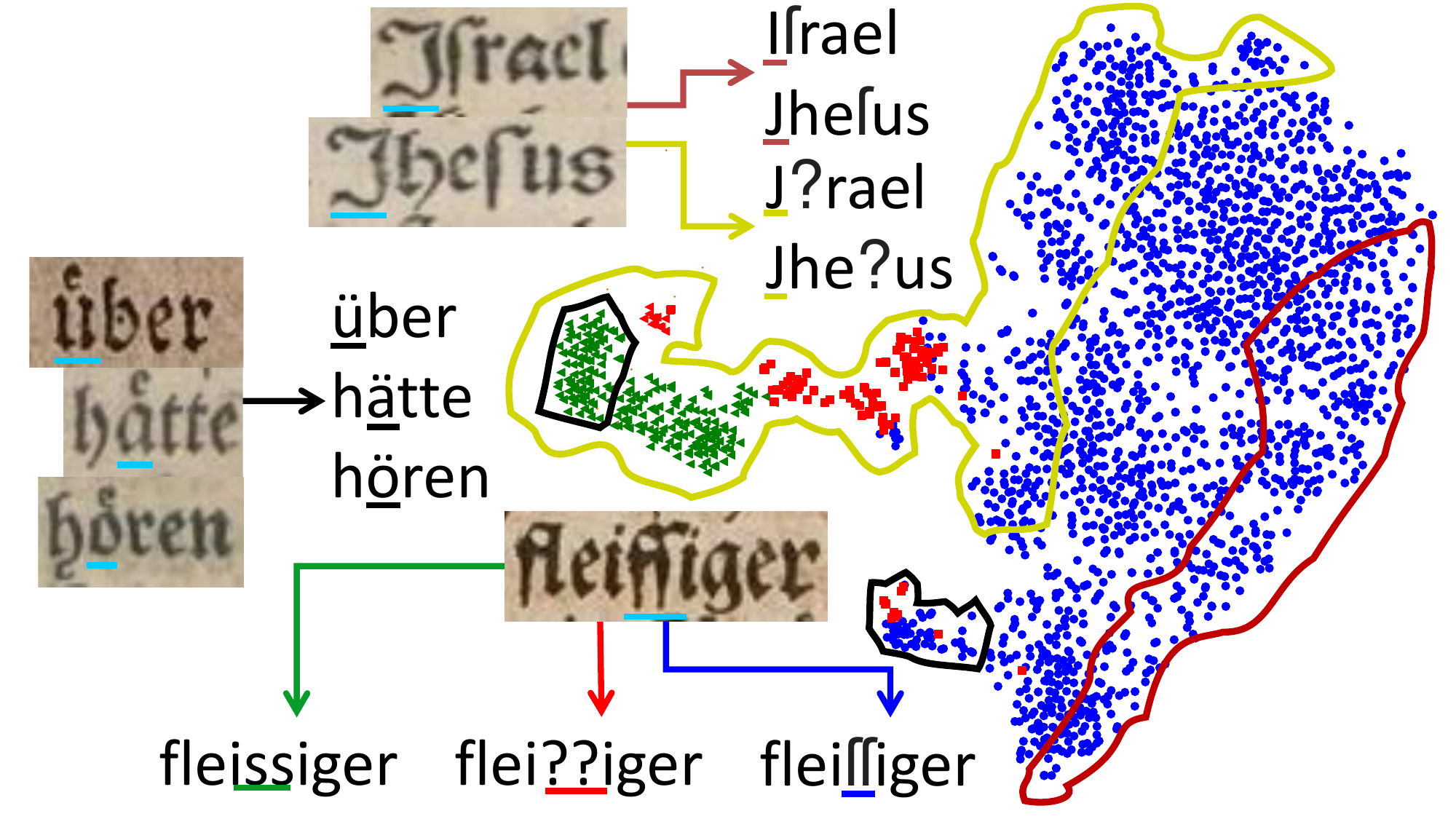}   
    \caption{Learned transcription style embeddings projected to 2D by t-SNE using output text edit distances on DTA TST-HO.
    Colors represent different transcription styles for certain characters. 
    Blue and green dots differentiate between modern and historic transcriptions of \textquotesingle{}long s\textquotesingle{}.
    The black polygons represent transcription styles that transcribe historic umlauts by their modern forms. 
    Embeddings inside the brown polygon distinguish between I and J, while those inside the yellow polygon do not.
    Embeddings producing inconsistent transcriptions are shown in red color.
    }
    \label{fig:experiments:real:TSNE}
\end{figure}

\paragraph{Structure of learned transcription style embeddings.}
Without a surprise, we observed that more similar transcription style embeddings produce more similar outputs. 
The Euclidean distances between embeddings and Levenshtein distances between the respective outputs are strongly correlated. 
We measured an almost perfect correlation for training documents with a high number of text lines. The correlation decreases with a lower number of training lines, but it plateaus approximately at 800 training lines and the Pearson correlation coefficient does not drop below 0.55.

Figure~\ref{fig:experiments:real:TSNE} shows a t-SNE projection 
of all learned transcription style embeddings. 
The distance metric we used for the projection is an average Levenshtein distance of respective output transcriptions on the DTA held out set TST-HO. 
We also categorized the embeddings based on how they transcribe \textquotesingle{}long s\textquotesingle{}, historic umlauted vowels and \textquotesingle{}J\textquotesingle{}. We were able to do this categorization automatically thanks to the grapheme-accurate transcriptions of the TST-C dataset.
The categories form a meaningful structure in the projection.

We also found embeddings that produced inconsistent transcriptions (red points in Figure~\ref{fig:experiments:real:TSNE}). 
Closer examination revealed that these embeddings represent documents with a high amount of annotation errors.
Interestingly, most of these embedding are grouped together.

\begin{figure}[t]
    \centering
    \includegraphics[width=\textwidth, trim=0mm 0mm 0mm 18mm, clip]{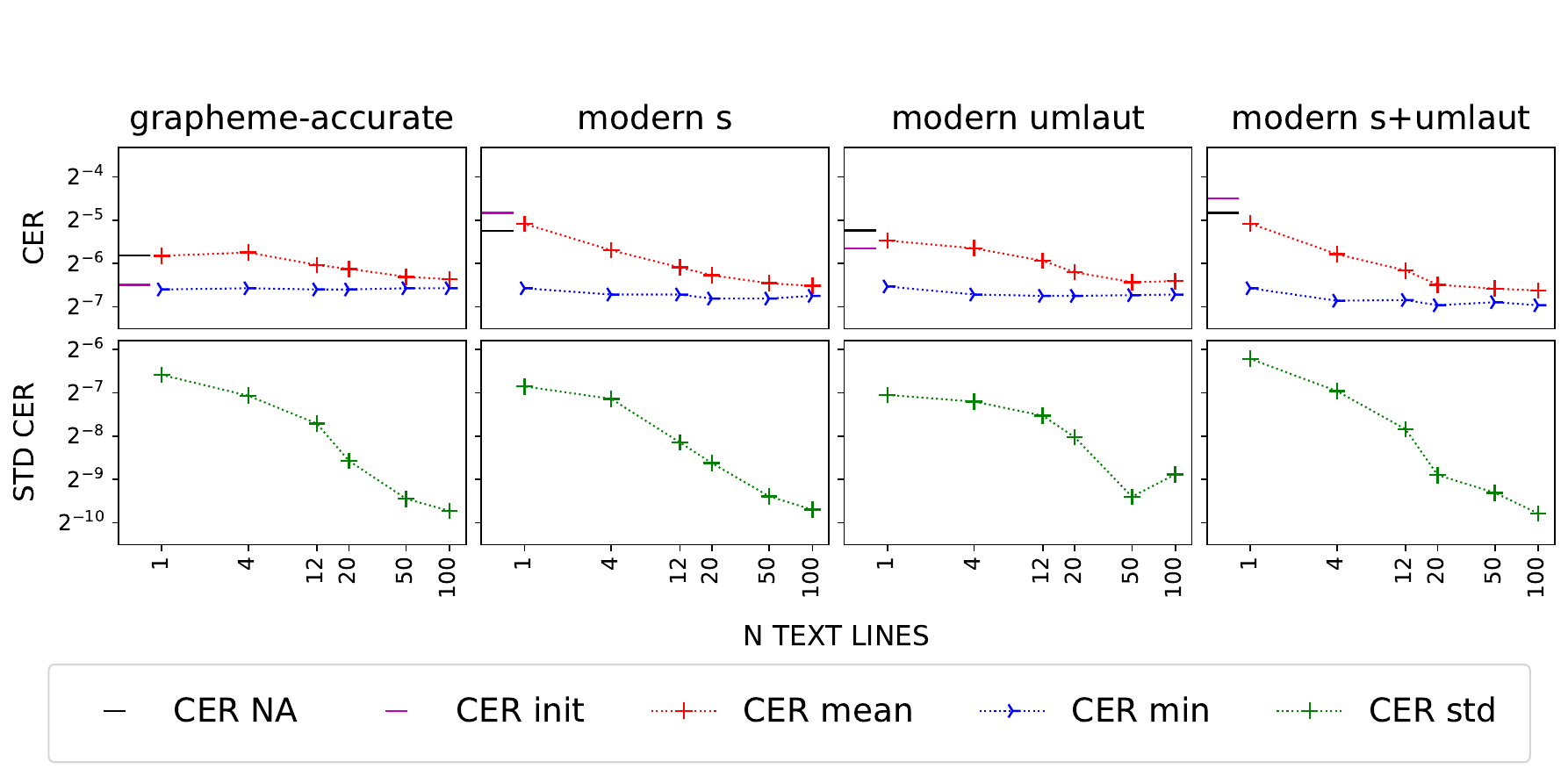}
    \caption{Adaptation of transcription style embeddings on unseen documents with LBFGS optimization. 
    Mean (red), standard deviation (green), and minimum (blue) test character error rate (CER) of the proposed system with adapted embeddings over 50 randomized text line sets.
    Black line marks the mean CER of the baseline network and the purple line marks the mean CER of our system with the initial embedding.
    The horizontal axis represents the number of text lines used for adaptation.
    Columns represent four different transcription styles.}
    \label{fig:LBFGS}
\end{figure}

\paragraph{Adaptation to a new transcription style.}
The proposed system provides two possibilities of how to choose a transcription style for users.
A straightforward way is to provide transcription style prototypes with descriptions (e.g. by clustering learned embeddings and manually creating their text descriptions). 
Alternatively, it is possible to estimate a transcription style embedding from a small set of exemplary text line transcriptions.
The transcription style embedding can be estimated by similar optimization as when training the whole network. 
The loss function is the same, but the network remains fixed and only the single embedding is optimized.
We use the LBFGS optimization algorithm which tries to better approximate local loss function curvature and it generally converges faster on problems with a low number of optimized variables.
We fixed the number of LBFGS iterations to 150 and we initialized the new embeddings as the mean of all training document embeddings.
The optimalization takes approximately 1 minute on GeForce RTX 2080 Ti.
We augmented the exemplary text lines in the same way as during full network training and did not use any additional regularization. 
The augmentation was necessary to avoid overfitting on small sets of exemplary text lines. 

We tested the transcription style adaptation on held-out dataset TST-C for which we generated three additional transcription styles from the grapheme-accurate transcriptions.
The four transcription styles are grapheme-accurate, modern s, modern umlaut, modern s+umlaut, where modern s replaces all \textquotesingle{}long s\textquotesingle{} characters with the modern \textquotesingle{}s\textquotesingle{} character and normal umlaut replaces the historic umlauts with their respective modern variants.

We repeated all optimizations $50\times$ on randomly sampled text lines from TST-C. 
We tested how the adaptation behaves with 1, 4, 12, 20, 50, and 100 exemplar text lines.
Figure~\ref{fig:LBFGS} shows that adaptation quality stabilizes and plateaus with the randomly sampled 100 exemplar text lines. 
In real use-cases, it is reasonable to expect that even fewer text lines would be needed as a user would probably pick representative text lines instead of sampling randomly (i.e. text lines which contain characters defining the transcription style). 
This expectation is supported by the fact that the adaptation achieves the same quality regardless of the number of exemplar lines for specific line sets (represented by the minimal error rates over the 50 runs).

Transcription accuracy after the adaptation is consistently better compared to the transcription style agnostic baseline network (NA). 
The reaching best relative improvements 42\%, 66\%, 65\%, and 77\% on the individual transcription styles.
Interestingly, the initial mean embedding produces the grapheme-accurate transcriptions and it provides comparable accuracy as the baseline network on the other styles.
The adaptations are able to achieve the same accuracy regardless of the target transcription style.

\section{Conclusion}

We proposed the Transcription Style Block (TSB) to solve the problems caused by inconsistent transcription styles present in real-world large-scale training datasets, and to allow OCR system users to easily choose their preferred transcription style.
By connecting TSB near the end of any standard text recognition network, the network learns to output transcriptions in different styles from the data, without any explicit knowledge about the transcription rules. 
We showed that our approach can handle extreme synthetic scenarios, where the transcription styles are random permutations of the character set, without any degradation of text recognition.
Our approach outperforms transcription style agnostic networks on a real-world large-scale dataset by a large margin.
The adaptation to a new transcription style is fast and can be done with a couple of representative text line transcriptions.

\subsubsection*{Acknowledgment.} This work has been supported by the Ministry of Culture Czech Republic in NAKI II project PERO (DG18P02OVV055).


\bibliographystyle{splncs04}
\bibliography{mybibliography}

\end{document}